\pdfoutput=1

\documentclass[11pt]{article}

\usepackage[final]{acl}

\usepackage{times}
\usepackage{latexsym}

\usepackage[T1]{fontenc}

\usepackage[utf8]{inputenc}

\usepackage{microtype}

\usepackage{inconsolata}

\usepackage{graphicx}

\usepackage{multirow}
\usepackage{enumitem}
\usepackage{bbm}
\usepackage{amsfonts}
\usepackage{subcaption}
\usepackage[linesnumbered,ruled,vlined]{algorithm2e}
\usepackage{tcolorbox}

\title{Adaptive Rectification Sampling for Test-Time Compute Scaling}

 \author{
 Zhendong Tan,\, Xingjun Zhang, \, Chaoyi Hu,\, Yancheng Pan, \, Shaoxun Wang \\
 School of Computer Science and Technology, Xi'an Jiaotong University \\ 
 \texttt{772316639@stu.xjtu.edu.cn}, \texttt{xjzhang@xjtu.edu.cn}
 }

\begin{document}
\maketitle
\begin{abstract}
The newly released OpenAI-o1 and DeepSeek-R1 have demonstrated that test-time scaling can significantly improve model performance, especially in complex tasks such as logical reasoning. Common test-time scaling methods involve generating more chains of thought (CoTs) or longer CoTs with self-correction.
However, while self-correction can improve performance, it may lead to significant token waste and reduce readability of the CoT if the reasoning steps are already correct. 
To demonstrate that large language models (LLMs) can rectify errors at a more fine-grained level, we propose \underline{A}daptive \underline{R}ectification Sampling (AR-Sampling), which can guide the LLMs to self-correction at the appropriate steps. AR-Sampling leverages a process-supervised reward model (PRM) as a verifier and constructed trigger sentences to guide the model in adaptive step-level rethinking.
Through the experiments on GSM8K and MATH500, it indicates that our approach enables the models to rethink in more fine-grained level, improving the accuracy of solutions, while generating a reasonable number of additional tokens. Our code is available at: \url{https://github.com/TanZhendong/AR-Sampling}.

\end{abstract}

\section{Introduction}
The newly released OpenAI-o1 and DeepSeek-R1 \citep{openaio1, deepseekr1} models have demonstrated remarkable capabilities in complex tasks such as logical reasoning \citep{deepseekmath, qwenmath} and code generation \citep{codellama}. With post-training techniques, represented by reinforcement learning, these models are capable of deep thinking, generating longer chain of thought (CoT) \citep{cot}, and improving the quality of their outputs. On the other hand, increasing the scale of pre-trained models requires more computational resources and data, which is difficult to afford. Therefore, many researchers are focusing on post-training and test-time scaling to enhance the model performance.

Common test-time scaling methods involve generating more CoTs, such as best of N, beam search, and other tree-of-thought approaches \citep{scalingttc, treeofthoughts, wan2024alphazero}. We refer to this method as increasing the \emph{width} of CoT, which means increasing the number of N or beams. Correspondingly, the o1-like model scales test-time inference by increasing the \emph{length} of CoT. By analyzing the deep thinking results of DeepSeek-R1, we can observe that during this phase, the model often produces phrases such as "Let me check again" or "Alternatively," leading to new solutions. This phenomenon is referred to as the "aha moment"\citep{deepseekr1}, which signifies allocating more thinking time to a problem by reevaluating its initial approach. However, this phenomenon typically arises spontaneously and uncontrollably. The model may still generate lengthy responses when problems are quite simple, a phenomenon known as "overthinking" \citep{chen2024not}.

Although the "aha moment" can enhance model performance, when the current reasoning steps are correct, checking and rethinking a new solution can result in a significant waste of tokens and reduce the readability of the CoT. Theoretically, if LLMs only rethink and rectify at the step where an error occurs, it could effectively reduce the number of tokens generated. However, during test-time, it is challenging to identify at which step the model made mistakes and to guide the model to regenerate from the incorrect step. As a result, the key research problem is: \emph{during test-time, how to guide LLMs to rethink at the appropriate moments?}

In this paper, we find that using a process-supervised reward model (PRM) as a verifier to check the reasoning steps can help identify potential errors. Consequently, we propose \underline{A}daptive \underline{R}ectification Sampling (AR-Sampling), which leverages the verifier and constructed trigger sentence to guide the model in adaptive step-level rethinking. On the one hand, AR-sampling can enhance the LLMs reasoning without generating redundant tokens; on the other hand, the sampled data can be used in other self-critique methods. At the same time, we demonstrate that LLMs have the ability to rethink at more fine-grained level, which is beneficial for addressing overthinking in the future.

\begin{figure*}[t]
  \includegraphics[width=\linewidth]{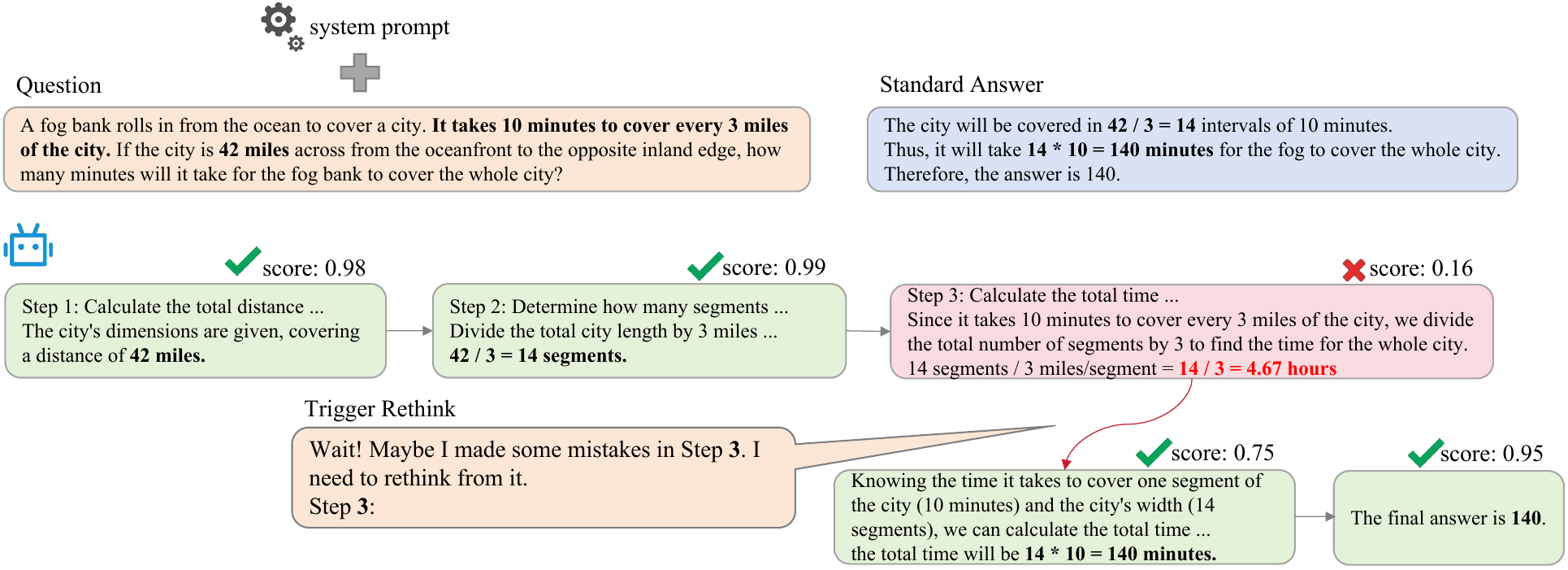}
  \caption {The framework of AR-Sampling. AR-Sampling uses PRM as a verifier to check each step. If the score is lower than the threshold, we consider this step unfavorable for the reasoning and use a trigger to force the model to rethink from the current step. By adjusting the threshold score and the maximum number of rethinks, we can dynamically control the generation budget.}
  \label{fig: framework}
\end{figure*}

The main contributions of this work are:
\begin{itemize}
	\item We propose AR-sampling, which utilizes PRM as a verifier to check the reasoning steps and use constructed trigger sentence to guide LLMs to rethink from the incorrect step.
	\item We demonstrate that LLMs have the ability to rethink at a more fine-grained level, which is beneficial for addressing overthinking.
	\item We evaluate our approach on the Llama3.2 and Qwen2.5 models. The results indicate that our approach enables the models to rethink in more fine-grained level, improving the accuracy of solutions, while increasing reasonable number of tokens.
\end{itemize}

\section{Related Work}
\paragraph{Test-Time Compute Scaling.}
\citet{scalingttc} provides a detailed demonstration of how LLMs can utilize additional computation at test time to improve accuracy. As illustrated in Section 1, the forms of test-time compute scaling can be categorized into increasing the \emph{width} or \emph{length} of the CoT. Increasing the width of CoT typically requires a verifier to aggregate or select the best answer from the proposer \citep{cobbe2021training}. If combined with majority voting \citep{selfconsistency}, the accuracy and stability of best-of-N sampling can be further improved. According to \citet{prm}, process-based verifier generally perform better than outcome-based verifier \citep{uesato2022solving}. Due to the branching nature of per-step predictions, we can also search within a tree-like solution space. Methods such as beam search and Monte Carlo tree search, which explore the tree of thoughts, can be more efficient and enable the model to perform better \citep{wan2024alphazero, treeofthoughts, xie2023decomposition}.

Increasing the \emph{length} of CoT always relies on the model's self-reflection capabilities, meaning the model can refine its own outputs, regardless of whether they are correct or not, to enhance its responses. \citet{selfrefine} demonstrates that LLMs can provide feedback and utilize it to self-refine. Building on this foundation, many applications leverage this self-reflection mechanism to improve the outputs of LLMs \citep{gou2023critic, selfbug}. In addition, manually inserting prompts can also trigger self-reflection \citep{chen2025iterative}. Beyond this, we can leverage reinforcement learning and direct preference optimization \citep{dpo} on sampled data for fine-tuning, enabling the model to achieve self-improvement \citep{qu2024recursive}.

Overall, according to \citet{liu2025can}, the compute-optimal test-time scaling strategy strongly depends on the choice of policy model, PRM, and problem difficulty. With an appropriate strategy, even smaller models can exceed much larger ones. Therefore, identifying the most efficient test-time scaling strategy is challenging and complex.

\paragraph{LLM Reasoning.}
LLM reasoning has always been an important research area. Its primary goal is to enhance the logical reasoning capabilities of LLMs, particularly in solving mathematical problems. CoT has become an essential process for LLMs to answer mathematical questions, as solving problems step by step can significantly improve the accuracy and increase the readability of the solving process, which is widely applied in recent works \citep{cobbe2021training, kojima2022large, cot, uesato2022solving}. Moreover, many test-time compute scaling methods also employ mathematical reasoning for validation \citep{chen2025iterative, scalingttc, beeching2024scalingtesttimecompute}.

\paragraph{Efficient Thinking.}
OpenAI-o1 and DeepSeek-R1 have already demonstrated the amazing potential that comes with deep thinking. However, they tend to generate a very large number of tokens in response, even when the questions are quite simple. The core objective of efficient thinking is to explore methods for scaling test-time compute efficiently and intelligently.

A commonly adopted approach is to adaptively set the width of the CoT. \citet{aggarwal2023let} and \citet{li2024escape} have explored the possibility of early stopping within self-consistency from different perspectives, aiming to prevent the model from excessive generation. Expanding upon this, \citet{wang2024make} incorporates prior knowledge about question difficulty to adaptively allocate inference resources. What's more, recent works focus on reducing the length of CoT, including token-budget-aware inference \citep{han2024token, yu2025think} and CoT compression \citep{luo2025o1pruner, ma2025cotV}. The aforementioned methods primarily focus on saving inference budget at the solution level. Instead, we concentrate on a more fine-grained step-level rethinking, aiming to further explore the self-correction mechanisms to improve generation efficiency.

\section{Adaptive Rectification Sampling}
\subsection{Preliminaries}
Best-of-N sampling is one of the most commonly used methods for test-time compute scaling. To better understand our method, we will first introduce the details of it.

Intuitively, when adopting best-of-N sampling, the model needs to generate N different samples through stochastic decoding methods such as top-k, top-p, and temperature sampling, and then select the one with the highest score from these candidates. When using an outcome-supervised reward model (ORM), a single score is assigned to a solution path. However, when using a PRM, since the model provides scores at each step, we need to aggregate these scores. Typically, we can use reduction operations $f(\cdot)$ to obtain the aggregated score, such as taking the minimum of these step scores, the product of them, or simply using the score from the final step.

After obtaining the scores for each solution path, we can select the final answer $a$ with the highest score from the answer set $A$:
\begin{equation}
  \label{eq: best-of-N}
  a_{BoN}=\mathop{\arg\max}\limits_{a\in A} f(r_a^{(1)}, r_a^{(2)},\dots,r_a^{(k)})
\end{equation}
$r_a^{(i)}$ represents the score of the i-th step in the solution path where the final answer is $a$.
We assume that there are $k$ steps for a solution.

Moreover, according to \citet{li2022making}, we can further improve the stability and accuracy of best-of-N sampling by utilizing majority voting, using weighted scores for selection:
\begin{equation}
  \label{eq: weighted best-of-N}
  a_{BoN}=\mathop{\arg\max}\limits_{a\in A} \sum_{i=1}^N \mathbbm{1}(a_i=a)f(r_{a_i}^{(1)},\dots)
\end{equation}

\subsection{AR-Sampling}
Given a question, AR-Sampling requires the model to generate step by step. AR-Sampling uses a verifier to identify incorrect steps and then constructs trigger sentences to guide the model to rethink from it. Next, we describe it in more detail in section 3.2.1 and 3.2.2.

\subsubsection{Adaptive Step Detection}

\begin{figure*}[htbp]
    \begin{subfigure}{0.49\linewidth}
        \includegraphics[width=\linewidth]{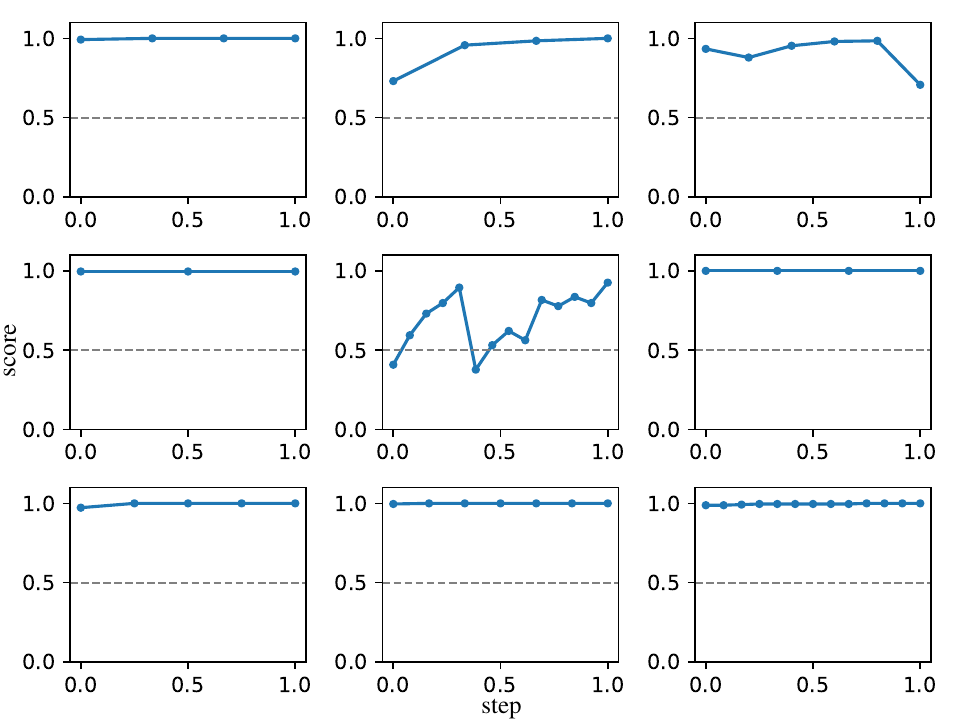}
        \caption{The final answer is correct.}
        \label{fig:correct_line}
    \end{subfigure}
    \hfill
    \begin{subfigure}{0.49\linewidth}
        \includegraphics[width=\linewidth]{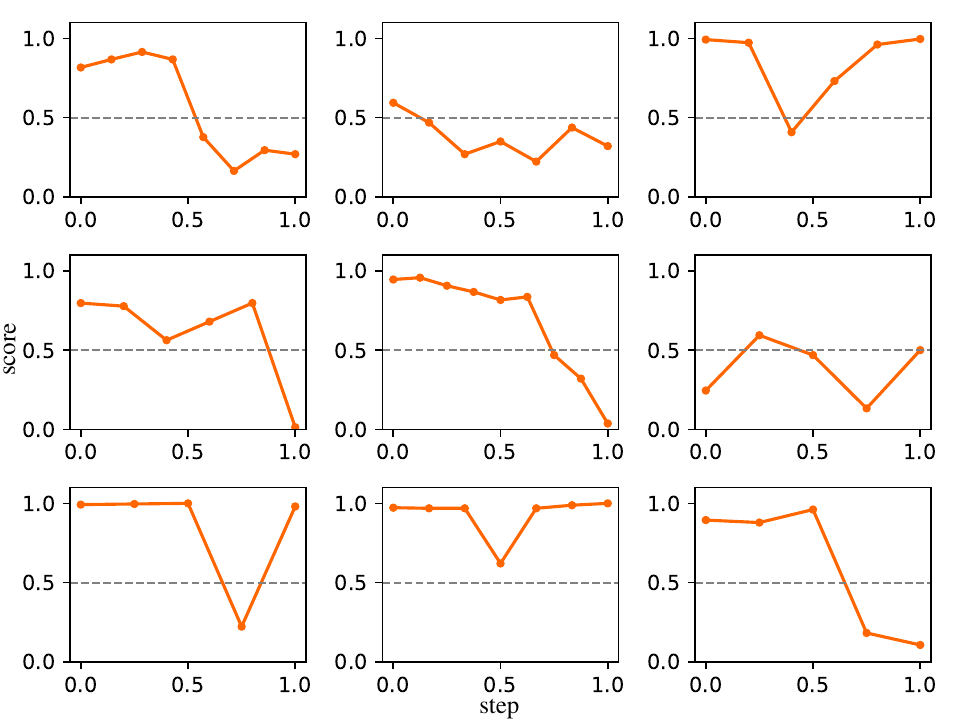}
        \caption{The final answer is wrong.}
        \label{fig:wrong_line}
    \end{subfigure}
    \caption{Samples of the PRM score. The x axis is the normalized step index. We classify them into correct (a) and wrong (b) according to the final answers. For the wrong cases, there is greater fluctuation, and the scores of some steps are very low. We believe that these steps are more likely to lead to wrong answers.}
    \label{fig:prm_scores}
\end{figure*}

We use PRM as a verifier to check each step. Typically, PRM is a LLM fine-tuned on datasets that are either manually annotated \citep{prm} or automatically annotated \citep{wang2023math}. It treats verification as a classification problem, where each step can be categorized into two or three classes: \emph{good}, \emph{bad}, and \emph{neutral}. \emph{Neutral} indicates that the step is correct but irrelevant to the reasoning goal. And it can be considered incorrect in two classes case. According to \citet{wang2023math}, there is not much difference between the binary and the three classification models. 

For the model architecture, while keeping the base model unchanged, we can use a token or sequence classification output head to replace the causal language model head. For the sequence classification model, each step is treated as a token sequence. For the token classification model, we can use the class of the last token as the class for the step \citep{vonwerra2022trl}. In addition, we can also utilize the original causal language model head, employing certain special tokens (such as '+' and '-') as markers for classes \citep{dong2024rlhf}.

Given a question $q$, the reward $r$ of the step $s_i$ can be considered as the probability of the good class:
\begin{equation}
  \label{eq: reward}
  r^{(i)}_a = \mathbb{P}(+|q, s_1, s_2, \dots, s_i)
\end{equation}

The score of each step can effectively reflect the correctness of it. Meanwhile, due to the close dependence among the steps, a wrong step is more likely to lead to a wrong answer. In order to display the distribution of the scores intuitively, we sampled several PRM scores during reasoning and plotted the trends in Figure \ref{fig:prm_scores}. 

We divide the scores into two groups based on whether the final answer is correct or not. For most of the correct cases, the scores of every step are relatively high and close to 1. For the wrong cases, there is greater fluctuation, and the scores of some steps are very low. We believe that these steps are more likely to lead to wrong answers. Therefore, we need to introduce triggers after these steps to guide the LLMs to rethink from them. Specifically, we introduced a threshold $p$, which ranges between 0 and 1. If the score of the current step is less than $p$, a trigger will be introduced; otherwise, the reasoning will continue. Generally, the larger the value of $p$ is, the more likely the model is to trigger the rethink, and the more tokens will be generated.

\subsubsection{Step-level Rectification}
When LLMs generate the solution step by step, they often use the word "Step" or similar tags as markers at the beginning of each step. Additionally, separators such as "\verb|\|n" or "\verb|\|n\verb|\|n" will be used to indicate the end of a step.
By setting the separator as a stop word, we can ensure that the model stops after each step of generation. After we identify steps to rethink, we can construct a trigger sentence to guide the LLM to conduct step-level rethink.

Figure \ref{fig: framework} provides a specific example to illustrate this process. In the system prompt, we will provide the information about the separator and the step marker. Then, we can parse the step index (in Figure \ref{fig: framework}, the step index is 3.) and construct a specific trigger. In order to ensure that the model follows the instructions of the rethink trigger, we also add the step marker to the trigger sentence. 

After the model generates a new solution step, we will continue to use the verifier to check it. It should be noted that sometimes the score of the rethink step is still lower than the threshold $p$ we set. This may be caused by reward hacking, problem difficulty, and the capabilities of the model. In order to prevent the model from repeatedly thinking about the same step, we set a maximum number $m$ of rethink attempts for single step. If the number of rethinking attempts exceeds $m$, we will no longer add the trigger for this step. We provide the detailed description of AR-Sampling for one time in Algorithm \ref{alg: ARS}.

\IncMargin{1em}
\begin{algorithm}[t]
	\SetKwInOut{Input}{Input}\SetKwInOut{Output}{Output}
	\SetKwFunction{AddTrigger}{AddTrigger}
	\Input{$llm$, $prm$, question $q$, threshold $p$, max attempts $m$, max length $l$, seperator $sep$} 
	\Output{model generation $s$}
	\BlankLine 
	
	$index$ = 1, $s$ = $q$, $count$ = 0\;
	\For{index $\leq$ $l$}{
		\If{index < $l$}{
			\tcp{set stop word}
			$step$ = $llm$.generate($s$, stop=$sep$)\;
			$score$ = $prm$.score($s$, $step$, $sep$)\;
			\If{score < $p$ \textbf{and} count < m }{
				$step$ = \AddTrigger{$step$}\;
				$count$ += 1\;
			}
			\Else{
				$count$ = 0\;
			}
		}
		\Else{
			$step$ = $llm$.generate($s$)
		}
		$s$ += $step$\;
		$index$ += 1\;
	}
	\textbf{return} $s$\; 

 	\caption{AR-Sampling for one time}
	\label{alg: ARS} 
\end{algorithm}
\DecMargin{1em} 

\subsection{Relationship with Other Sampling Methods}
In section 3.2, we introduce AR-Sampling and provide a detailed description of a single generation instance. Our approach emphasizes step detection and prompt triggering, which is orthogonal to several generation methods. When employing AR-Sampling during test-time, it can be combined with best-of-N sampling to easily achieve scaling. 

To better understand AR-Sampling, we compare it with the most commonly used methods, best-of-N and beam search, as follows:
\begin{itemize}
	\item \textbf{Best-of-N:} Best-of-N can be regarded as a special case of the AR-Sampling algorithm with $p=0$, where no rethinking is performed. In this setting, each solution path includes all tokens generated by the model, including any incorrect steps before rethinking.
	\item \textbf{Beam search:} Unlike beam search, which keeps only the highest-scoring steps, AR-Sampling preserves all intermediate steps. This allows the model to repair errors later, making the process an explicit form of self-correction.
\end{itemize}

\section{Experiments}

\begin{table*}[t]
\small
\centering
\begin{tabular}{ccccccc}
\hline
\textbf{Model} (-Instruct) & \textbf{Method} & \textbf{Pass@2} & \textbf{Pass@4} & \textbf{Pass@8} & \textbf{Pass@16} & \textbf{Pass@32} \\
\hline

 \multirow{5}{*}{Llama3.2-1B} & SC & 44.9 & 54.0 & 59.2 & 63.8 & 66.8\\  
 & BoN & 55.3 & 62.5 & 67.1 & 70.3 & 71.8\\  
 & \textbf{AR+BoN} & \textbf{58.4} & \textbf{65.4} & \textbf{71.5} & \textbf{74.4} & \textbf{74.5}\\
 & BoN+SC & 55.3 & 61.0 & 64.7 & 68.0 & 70.3   \\
 & \textbf{AR+BoN+SC} & \textbf{58.4} & 65.0 & 70.4 & 73.7 & 73.7 \\
 \hline

 \multirow{5}{*}{Llama3.2-3B} & SC & 78.8 & 85.2 & 87.7 & 88.8 & 88.6\\  
 & BoN & 85.0 & 86.8 & 88.0 & 88.9 & 89.8\\  
 & \textbf{AR+BoN} &  \textbf{86.1} & \textbf{88.4} & \textbf{89.5} & \textbf{90.1} & \textbf{90.4}\\
 & BoN+SC & 85.0 & 87.1 & 88.6 & 89.6 & 89.5\\
 & \textbf{AR+BoN+SC} & \textbf{86.1} & 88.3 & 89.3 & 89.8 & 90.3\\
 \hline
 
  \multirow{5}{*}{Qwen2.5-7B} & SC & 87.6 & 92.0 & 92.3 & 92.9 & 93.5\\  
 & BoN & \textbf{91.1} & 92.3 & 93.2 & 93.3 & 93.7\\  
 & \textbf{AR+BoN} &  90.1 & \textbf{92.5} & 93.1 & \textbf{93.6} & \textbf{94.2}\\
 & BoN+SC & \textbf{91.1} & 92.3 & 93.2 & 93.3 & 93.7\\
 & \textbf{AR+BoN+SC} & 90.1 & 92.3 & \textbf{93.3} & 93.5 & 93.7\\
 \hline
 
\end{tabular}
\caption{AR-Sampling can improve the accuracy (\%) on GSM8K.}
\label{tab:gsm8k}
\end{table*}

\subsection{Experiment Setup}
\paragraph{Models and Datasets.}
In our experiments, we evaluate AR-Sampling in mathematical reasoning. For the proposer model, we choose the Llama3.2-1B-Instruct, Llama3.2-3B-Instruct \citep{grattafiori2024llama}, and Qwen2.5-7B-Instruct models \citep{yang2024qwen2}. To ensure the models have sufficient instruction-following capabilities, we use the instruction-tuned versions rather than the base models. For the verifier model, we choose the PRM trained from Llama3.1-8B-Instruct on RLHFlow/Deepseek-PRM-Data \citep{xiong2024rlhflowmath}. The verifier use '+' and '-' tokens to label the class of a step, as illustrated in section 3.2.1. A larger model size ensures that the verifier can effectively identify potential wrong steps for AR-Sampling and efficiently select the best answer.
For datasets, we select two representative datasets: GSM8K \citep{cobbe2021training} and MATH500 \citep{prm}.

\paragraph{Baselines.}
We combine best-of-N with AR-Sampling to achieve test-time scaling. Because with best-of-N sampling, all the solution steps are retained, which facilitates our analysis of the solution path and allows us to explore the step-level rectification capabilities of the model. Consequently, we compare our approach against best-of-N sampling and its variants:
\begin{itemize}
	\item \textbf{Self Consistency (SC):} SC selects the final answer using a majority voting mechanism \citep{selfconsistency} and does not require a verifier. Comparing with SC can demonstrate the effectiveness of the verifier.
	\item \textbf{Best-of-N (BoN):} BoN selects the best answer with the highest score, as illustrated in Equation (\ref{eq: best-of-N}). Additionally, according to \citet{li2022making}, we can combine BoN with majority voting, as illustrated in Equation (\ref{eq: weighted best-of-N}), which we denote as BoN+SC.
\end{itemize}

\paragraph{Metrics.}
The primary metric is the accuracy of the final answer. When scaling at test-time, we use \emph{pass@N} to represent the accuracy when the model generate N samples. A key issue when evaluating the answers is that there are many equivalent expressions in mathematics, such as 1/2 and 0.5. The standard approach \citep{lewkowycz2022solving} to address it is to let the model generate answers in LaTeX format and use symbolic computation to verify whether they are equivalent.

\paragraph{Implementation.}
We run the experiments on a single NVIDIA A100 (80GB) GPU. For the proposer model, we use the vLLM inference engine \citep{kwon2023efficient}, and for the verifier, we use Hugging Face Transformers \citep{wolf-etal-2020-transformers}. 
For the parameters, we set \( m = 1 \) and adjust \( p \) to control the rethinking. Generally speaking, the stronger the model's capability or the easier the dataset difficulty, the larger \( p \) becomes. 
Consequently, on GSM8K, we set \( p \) to 0.6, 0.7, and 0.8 for the 1B, 3B, and 7B models, respectively. On MATH500, we set them to 0.3, 0.5, and 0.5.
Additionally, we set the maximum value of $N$ to 32, which is sufficient to reflect the effect of test-time scaling.
We set the stop word to "\verb|\|n\verb|\|n", and the system prompt is provided in Appendix \ref{sec: system prompt}.

\subsection{Results}
We evaluate AR-Sampling from three aspects: the accuracy of test-time scaling, step-level rethinking efficiency, and ablation study on parameters $p$ and $m$. The details are as follows.

\subsubsection{Accuracy}

\paragraph{GSM8K.}
GSM8K (Grade School Math 8K) is a dataset comprising high-quality, linguistically diverse grade school math word problems. We use the test dataset containing 1.32K problems. While the questions are straightforward for humans, they effectively evaluate the multi-step reasoning capability of the model. 

The results are presented in Table \ref{tab:gsm8k}. As $N$ increases, the accuracy of all approaches shows significant improvement. With only a marginal improvement observed from \emph{pass@16} to \emph{pass@32}, it suggests that the performance is almost convergence. Additionally, it can be confirmed that the verifier model effectively improves the accuracy compared with SC.
We observe that AR-Sampling can improve the accuracy on GSM8K in almost all cases. This indicates that the model can use the self-correction mechanism to improve its performance. 
What's more, we note that in some cases, when combined with major voting, the accuracy will slightly decrease, which may be caused by bias or hallucination. 
Finally, for Qwen2.5-7B-Instruct, since its performance on GSM8K is saturated and it is out-of-distribution with the verifier, the results of SC are comparable to those of BoN. However, using AR-Sampling still brings improvement, indicating that our hypothesis generalizes to some degree.

\begin{table*}[t]
\small
  \centering
\begin{tabular}{ccccccc}
\hline
\textbf{Model} (-Instruct) & \textbf{Method} & \textbf{Pass@2} & \textbf{Pass@4} & \textbf{Pass@8} & \textbf{Pass@16} & \textbf{Pass@32} \\
\hline

 \multirow{5}{*}{Llama3.2-1B} & SC & 28.2 & 34.4 & 37.6 & 40.8 & 43.8\\  
 & BoN & 31.8 & 36.6 & 39.6 & 42.2 & 41.8\\  
 & \textbf{AR+BoN} & \textbf{32.4} & 37.6 & 41.8 & 43.4 & 44.0\\
 & BoN+SC & 31.8 & \textbf{38.6} & 41.4 & 44.4 & 45.0   \\
 & \textbf{AR+BoN+SC} & \textbf{32.4} & 36.8 & \textbf{42.2} & \textbf{46.0} & \textbf{47.8} \\
 \hline

 \multirow{5}{*}{Llama3.2-3B} & SC & 46.8 & 53.4 & 56.4 & 59.0 & 61.0\\  
 & BoN & \textbf{49.6} & 53.2 & 54.8 & 57.2 & 55.8\\  
 & \textbf{AR+BoN} &  47.2 & 51.6 & 52.6 & 57.0 & 56.6\\
 & BoN+SC & \textbf{49.6} & \textbf{55.4} & \textbf{57.0} & 60.0 & 62.0\\
 & \textbf{AR+BoN+SC} & 47.2 & 54.4 & 56.6 & \textbf{62.0} & \textbf{63.2}\\
 \hline

 \multirow{5}{*}{Qwen2.5-7B} & SC & 66.8 & 73.4 & 77.2 & 79.2 & 80.0\\  
 & BoN & 70.2 & 72.6 & 71.0 & 70.6 & 71.2\\  
 & \textbf{AR+BoN} &  \textbf{70.4} & 70.2 & 72.6 & 73.4 & 73.4\\
 & BoN+SC & 70.2 & \textbf{74.8} & \textbf{77.8} & 79.0 & 79.8\\
 & \textbf{AR+BoN+SC} & \textbf{70.4} & 73.6 & 76.8 & \textbf{79.6} & \textbf{81.0}\\
 \hline
  
\end{tabular}
  \caption{AR-Sampling can improve the accuracy (\%) on MATH500.}
\label{tab:math500}
\end{table*}

\begin{table}[t]
\small
  \centering
\begin{tabular}{cccc}
\hline
\textbf{Model} & \textbf{Method} & \textbf{GSM8K} & \textbf{Math500}  \\
\hline

R1-Distilled & - & 377.9 & 1113.3 \\
 \hline

 \multirow{2}{*}{Llama3.2-1B} & BoN & 214.9 & 566.1 \\  
 & AR & 391.8 & 986.6 \\  
 \hline

 \multirow{2}{*}{Llama3.2-3B} & BoN & 208.0 & 478.7\\  
 & AR & 254.1 & 858.7\\ 
 \hline

 \multirow{2}{*}{Qwen2.5-7B} & BoN & 195.3 & 395.6\\  
 & AR & 210.2 & 461.6\\  
 \hline
 
\end{tabular}
\caption{The average number of tokens per solution.}
\label{tab:tokens}
\end{table}

\paragraph{MATH500.}
To avoid over-fitting, \citet{prm} expanded the PRM training set to include part of the MATH test problems \citep{hendrycks2021measuring}. Therefore, they selected 500 test problems for evaluation, referred to as MATH500. This dataset includes knowledge areas such as precalculus and algebra, which are challenging for LLMs. We believe it can effectively reflect the reasoning ability of LLMs.

The results are shown in Table \ref{tab:math500}. We observe that the AR-Sampling can improve the accuracy across most scenarios.  
However, for Llama3.2-3B-Instruct and Qwen2.5-7B-Instruct, it is interesting that the accuracy of AR-Sampling decreases when $N=4, 8$. We believe this is due to out-of-distribution, where PRM cannot effectively guide the LLMs. In the case of Qwen, the performance of SC even surpasses that of using PRM. 
Another reason is that because the question is hard, the self-correction mechanism cannot consistently improve performance—in fact, the proposer model may even change a correct step to an incorrect one after rethinking.

Moreover, we find that applying AR-Sampling to stronger models poses new challenges. As models become more capable, their reasoning is already more reliable, and the PRM must provide more precise guidance to be effective. Our focus here is to establish the feasibility of fine-grained step-level self-correction, particularly at moderate scales where errors are more frequent. Exploring joint scaling of both the PRM and generation model remains an important direction for future work.

\begin{figure}[t]
    \begin{subfigure}{0.49\linewidth}
        \includegraphics[width=\linewidth]{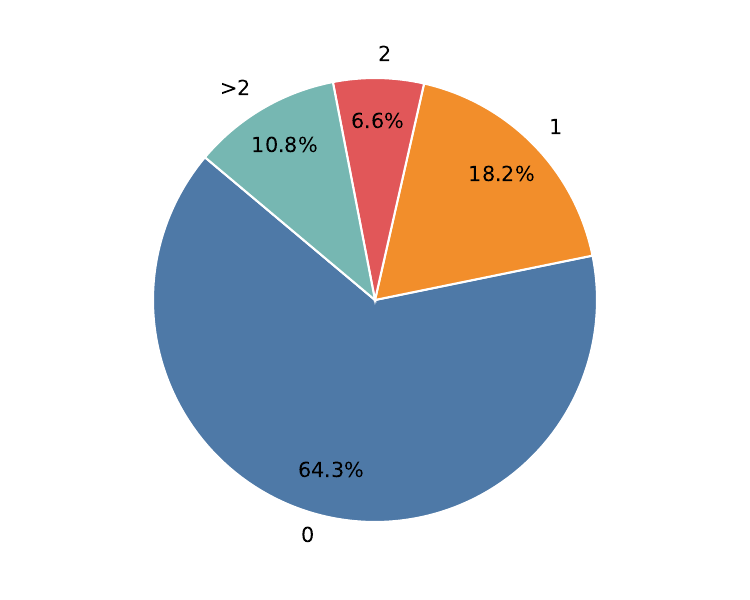}
        \caption{Distribution on GSM8K}
        \label{fig:rethink1}
    \end{subfigure}
    \hfill
    \begin{subfigure}{0.49\linewidth}
        \includegraphics[width=\linewidth]{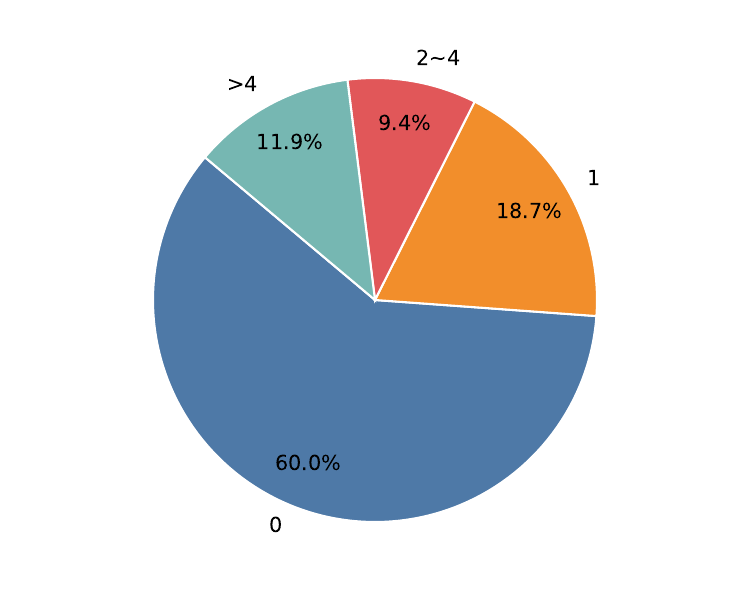}
        \caption{Distribution on MATH500}
        \label{fig:rethink2}
    \end{subfigure}
    \caption{The distribution of the number of rethinks for the Llama3.2-1B-Instruct model generation.}
    \label{fig:rethink}
\end{figure}

\begin{figure*}[t]
    \begin{subfigure}{0.49\linewidth}
        \includegraphics[width=\linewidth]{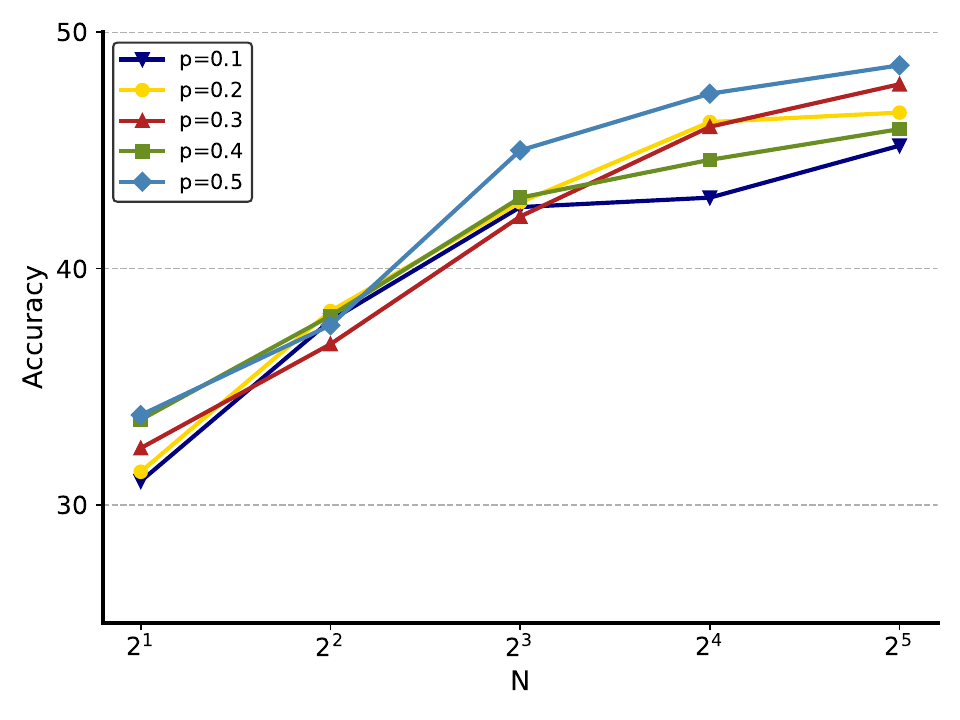}
        \caption{Ablation study on parameter $p$.}
        \label{fig:ablation_p}
    \end{subfigure}
    \hfill
    \begin{subfigure}{0.49\linewidth}
        \includegraphics[width=\linewidth]{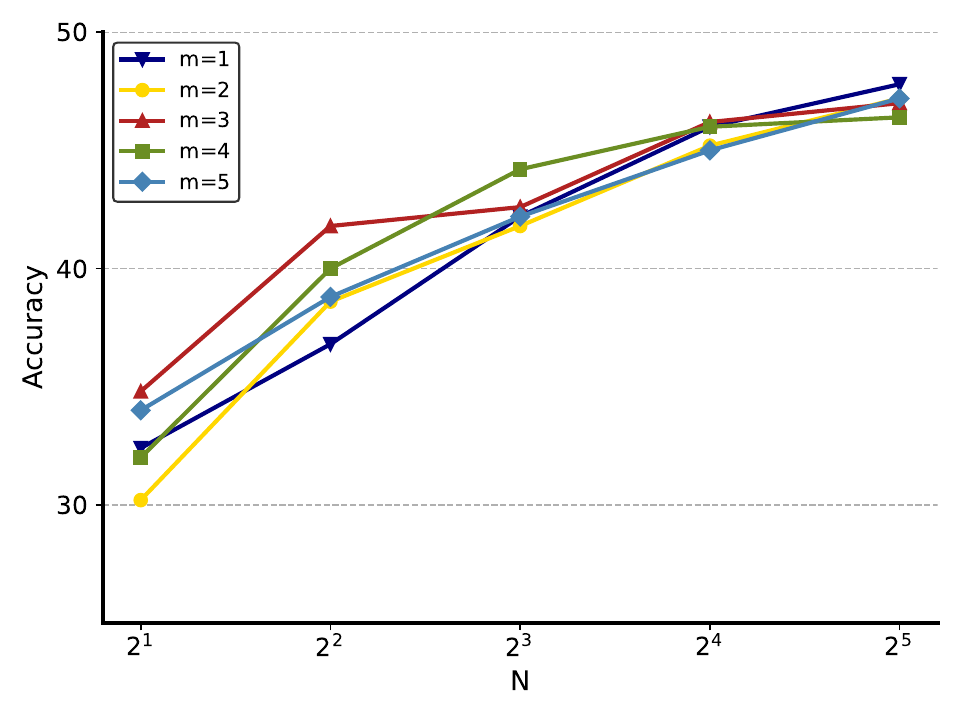}
        \caption{Ablation study on parameter $m$.}
        \label{fig:ablation_r}
    \end{subfigure}
    \caption{An ablation study on the Llama3.2-1B-Instruct model, exploring its performance on the MATH500 dataset under different parameters. The accuracy is calculated using major voting.}
    \label{fig:ablation}
\end{figure*}

\subsubsection{Rethinking Efficiency}
\paragraph{the Number of Tokens.}
To better understand the rethinking overhead of AR-Sampling, we measure the average number of tokens generated per solution. The results are shown in Table \ref{tab:tokens}.

We observe that as models scale up and their capabilities improve, the average number of tokens per solution gradually decreases. Larger models can solve problems with fewer steps and fewer tokens per step.
Additionally, the rethinking overhead diminishes with increased model size, allowing models to perform self-correction more efficiently.
We also measure the average number of tokens generated by DeepSeek-R1-Distilled-Qwen-7B, a model fine-tuned on DeepSeek-R1 data \citep{deepseekr1}. Since the model will think before generating solutions, the average number of tokens increases significantly, especially when the model repeatedly rethinks. In comparison, step-level rethinking obtain lower overhead.

\paragraph{Adaptive Rethink.}
We investigate the ratio of our trigger sentences in the total generated solutions. Figure \ref{fig:rethink} shows the distribution of the number of rethinking steps for the Llama3.2-1B-Instruct model on GSM8K and MATH500.

By controlling the threshold \( p \), we can adjust the proportion of rethink. For approximately 40\% of the solutions, trigger sentences are introduced. And for the remaining 60\%, since the PRM score is relatively high, it is unnecessary for the model to rethink.
From the difficulty perspective, the model needs less rethinking for simpler questions.
Additionally, about half of the rethinking solutions introduce the trigger sentence only once. This indicates that the model can effectively influence the correctness of subsequent steps through one critical self-correction.

\subsubsection{Ablation Study}
To explore the impact of parameters $p$ and $m$ in AR-Sampling, we conduct an ablation study on the Llama3.2-1B-Instruct model as an example. We vary $p$ from 0.1 to 0.5 and $m$ from 1 to 5. The results are presented in Figure \ref{fig:ablation}.

For the parameter \( p \), it cannot effectively introduce the trigger sentence when $p$ is too small. As a result, the improvement is not significant. Generally speaking, as the value of \( p \) increases, the number of rethinking iterations increases, leading to better performance. 
Additionally, when \( N \) is less than 16, the impact on accuracy is limited. This indicates that without fine-tuning, the self-correction of the model will be not precise, and a small number of generation cannot effectively improve performance.  

For the parameter \( m \), although increasing $ m $ introduces more rethinking iterations, setting it too high can cause overthinking at a step or even lead to cyclic generation. Therefore, we find that $ m = 1 $ is sufficient to guide the model toward effective self-correction.

From the perspective of generation length, as \( p \) and \( m \) increase, the length of the CoT grows rapidly. Therefore, we need to set appropriate values to ensure efficiency.  
Finally, different models and datasets require different parameter configurations. For easier datasets such as GSM8K, since the accuracy is already high, the PRM scores are typically high as well. Therefore, it is necessary to appropriately increase \( p \) or \( m \) to ensure the model can effectively use the self-correction mechanism.

\section{Conclusions}
In this work, we propose Adaptive Rectification Sampling (AR-Sampling), which leverages a verifier and constructed trigger sentence to guide the model in adaptive step-level rethinking. With more fine-grained rethinking, AR-Sampling can improve the accuracy of solutions at test-time while generating a reasonable number of additional tokens. Through our research, we demonstrate that LLMs have the ability to rethink at a more fine-grained level, which is beneficial for addressing overthinking in the future.  

\section*{Limitations}
Our proposed method has several limitations, which we believe can be addressed in future work.
First, as noted in DeepSeek-R1, the “aha moment” may emerge spontaneously when scaling reinforcement learning. If the goal is to mitigate overthinking, simply relying on trigger sentences at test time is insufficient. A more promising direction is to integrate fine-grained rectification into reinforcement learning itself, enabling better control of the “aha moment” phenomenon.
Second, while deep reasoning could potentially benefit other domains such as code generation, we did not validate our approach in these areas due to the lack of suitable PRMs and datasets.
Finally, our study has not yet provided a comprehensive scaling analysis. Extending step-level self-correction to stronger LLMs and developing more capable PRMs remain important directions for future work.
We hope that future work will further investigate these aspects and expand the applications of self-correction to enhance model performance.



\bibliography{custom}

\appendix

\section{System Prompt}
\label{sec: system prompt}

The system prompt used in our experiments is shown below:

\vspace{0.5em}

\begin{tcolorbox}[colback=gray!5,colframe=gray!50,title=System Prompt,width=0.9\textwidth]
Solve the following math problem efficiently and clearly:
\\

- For simple problems (2 steps or fewer):
    
    Provide a concise solution with minimal explanation.
\\

- For complex problems (3 steps or more):
    
    Use this step-by-step format:
\\
        
\#\# Step 1: [Concise description]

[Brief explanation and calculations]
\\
        
\#\# Step 2: [Concise description]

[Brief explanation and calculations]
\\

...
\\

Regardless of the approach, always conclude with:
\\

Therefore, the final answer is: \$$\backslash$$\backslash$boxed\{answer\}\$. I hope it is correct.
\\

Where [answer] is just the final number or expression that solves the problem.

\end{tcolorbox}

\section{Case Study}
We provide the case data for a specific problem in GSM8K. For AR-Sampling data, the model use step-level rethinking to rectify the wrong step. 
In contrast, DeepSeek-R1 performs solution-level verification and reflection, which may sometimes consume additional tokens unnecessarily on easy questions.
\\
\vspace{0.5em} 
\noindent
\newpage
\onecolumn

\begin{tcolorbox}[colback=gray!5,colframe=gray!50,title=Question,width=0.9\textwidth]
Every day, Wendi feeds each of her chickens three cups of mixed chicken feed, containing seeds, mealworms and vegetables to help keep them healthy.  She gives the chickens their feed in three separate meals. In the morning, she gives her flock of chickens 15 cups of feed.  In the afternoon, she gives her chickens another 25 cups of feed.  How many cups of feed does she need to give her chickens in the final meal of the day if the size of Wendi's flock is 20 chickens?
\end{tcolorbox}

\begin{tcolorbox}[colback=teal!5,colframe=teal!50,title=Answer,width=0.9\textwidth]
If each chicken eats 3 cups of feed per day, then for 20 chickens they would need 3*20=<<3*20=60>>60 cups of feed per day.

If she feeds the flock 15 cups of feed in the morning, and 25 cups in the afternoon, then the final meal would require 60-15-25=<<60-15-25=20>>20 cups of chicken feed.
\end{tcolorbox}

\begin{tcolorbox}[colback=blue!5,colframe=blue!50,title=DeepSeek-R1 Answer,width=0.9\textwidth]
Okay, let me try to figure out this problem. 

...

So, Wendi needs to give 20 cups in the final meal.
\\

\textcolor{red}{Wait, let me make sure I didn't make a mistake. Let me check again.} Each chicken has 3 cups per day. 20 chickens would need 3*20=60. Morning is 15, afternoon is 25. 15+25=40. 60-40=20. Yeah, that seems right. Hmm.
\\

\textcolor{red}{Alternatively, maybe there's another way to think about it.} Let's see. Maybe per meal, how much does each chicken get? Wait, but the problem says she gives them three separate meals, but doesn't specify if each meal is equal. So, in the morning, she gives 15 cups to the flock, afternoon 25 cups, and the final meal is the remaining. So, since total per day is 60, then 15 +25 + x =60. Solve for x: x=60-15-25=20. So, that's the same answer. So, 20 cups for the final meal.
\\

\textcolor{red}{Alternatively, maybe check per chicken.} Each chicken gets 3 cups a day. If the total feed given in the morning is 15 cups, then per chicken, morning meal is 15 cups divided by 20 chickens. Let me compute that: 15/20 = 0.75 cups per chicken in the morning. Then afternoon is 25 cups, so 25/20=1.25 cups per chicken in the afternoon. So, so far, each chicken has received 0.75 +1.25 =2 cups. Since they need 3 cups total, they need 1 more cup per chicken in the final meal. So, 1 cup per chicken times 20 chickens is 20 cups. So, that's the same answer.
\\

Therefore, both methods give the same result. So, Wendi needs to give 20 cups in the final meal. I think that's the correct answer. \textcolor{red}{Let me just check if there's any other way to interpret the problem.} ... So, the total has to be 60, so the remaining is 20. So, that seems correct. Therefore, I think 20 cups is the answer.
\\

**Final Answer**
\\

Wendi needs to give her chickens $\backslash$boxed\{20\} cups of feed in the final meal.
\end{tcolorbox}

\begin{tcolorbox}[colback=orange!5,colframe=orange!60,title=AR-Sampling Answer,width=0.9\textwidth]
\#\# Step 1: Calculate the total cups of feed given in the morning and afternoon meals.

Total cups given in morning and afternoon = 15 cups + 25 cups = 40 cups.
\\

\#\# Step 2: Calculate the remaining cups of feed needed for the final meal.

Remaining cups needed for the final meal = Total cups given - Total cups given in morning and afternoon = 40 cups - 40 cups = 0 cups.
\\

\textcolor{red}{Wait! Maybe I made some mistakes in Step 2. I need to rethink from it.}

\#\# Step 2:  Calculate the total cups of feed for the day.

Total cups of feed for the day = 3 meals * 20 chickens = 60 cups.
\\

\#\# Step 3: Calculate the remaining cups of feed needed for the final meal.

Remaining cups needed for the final meal = Total cups of feed for the day - Total cups given in morning and afternoon = 60 cups - 40 cups = 20 cups.
\\

The final answer is: \$$\backslash$boxed\{20\}\$
\end{tcolorbox}

\end{document}